\documentclass[12pt]{article}
\usepackage[utf8]{inputenc}
\usepackage{tikz}
\usetikzlibrary{arrows}
\usepackage{amssymb,amsthm,verbatim, fontenc,mathtools}
\usepackage[most]{tcolorbox}

\usepackage{amsmath,amsfonts,graphicx, bm}
\usepackage{vmargin, url}
\usepackage{setspace}
\usepackage[authoryear]{natbib}

\title{\vspace{-2.5cm}
Arbitrary-Length Generalization for Addition\\
in a Tiny Transformer\footnote{I would like to thank Fernanda Cristiane de Oliveira for helping me to make parts of this work clearer.}}
\author{Alexandre Galv\~ao Patriota\thanks{\footnotesize email: {\tt patriota@ime.usp.br}}\\
     {\it \footnotesize Departamento de Estat\'istica, IME,
     Universidade de S\~ao Paulo}
     \vspace{-0.1cm}\\
     {\it \footnotesize Rua do Mat\~ao, 1010, S\~ao Paulo/SP, 05508-090, Brazil}
         \\ 
}

\date{}
\begin{document}

\maketitle
\begin{abstract}
This paper introduces a novel training methodology that enables a  Transformer model to generalize the addition of two-digit numbers to numbers with unseen lengths of digits. The proposed approach employs an autoregressive generation technique, processing from right to left, which mimics a common manual method for adding large numbers. To the best of my knowledge, this methodology has not been previously explored in the literature. All results are reproducible, and the corresponding R code is available at: \url{https://github.com/AGPatriota/ALGA-R/}.

\end{abstract}

\section{Introduction}

The Transformer architecture, as introduced by \cite{Vaswani2017}, appears sufficiently robust to learn how to generalize addition, a fundamental operation (\verb|a+b=c|) taught in elementary school. However, \cite{Nogueira2021} demonstrated that Transformers struggle to generalize this simple procedure effectively. Although some researchers have explored the use of both simplified and complex scratchpads to aid in training Transformers \citep{Nyeetal2021, Leeetal2024}, they have not achieved generalization to numbers with arbitrary digit lengths. Recently, \cite{McLeishetal2024} argue that, by integrating an embedding for each digit that encodes its position relative to the start of the number, it is possible to train Transformers on 20-digit numbers and achieve approximately 99\% accuracy on addition problems involving up to 100 digits. However, the authors do not study the accuracy for numbers exceeding 100 digits, which leaves an open question about the scalability of this approach to even larger numbers. This gap presents a significant opportunity for future research to explore the limits of Transformer generalization in arithmetic operations.

This paper introduces a novel training methodology that, unlike previous efforts, considers only two-digit numbers and does not rely on scratchpads but instead uses specially designed training instances that adequately represent the generative process of addition. I tested the model on addition tasks involving integers up to one thousand digits, where it consistently achieved 100\% accuracy across all test scenarios. This perfect accuracy underscores the effectiveness of the approach, making traditional benchmarking comparisons redundant as they would not provide further insights into the model’s performance. This methodology could potentially be adapted for more complex arithmetic operations or mathematical theorems, offering new avenues for computational mathematics. In the Github repository, a \verb|R| script named \verb|Testing-Digits.R|  is available for readers to test the trained model. This script checks if the generated output is correct (i.e., the predicted addition matches the actual addition) and displays the respective numbers and results. Below is an example for 50-digit numbers, chosen at random using \verb|set.seed(10)|.
 
\begin{tcolorbox}[colback=lightgray, sharp corners, boxrule=0.5mm]
\begin{verbatim}
source('Testing-Digits.R')
 Digit #50 
[1] "The output is TRUE"
[1] "x=89675627969177656514819490691831725109908874980671"
[1] "y=32029996942446258125998499183326035828805968783222"
[1] "Real x+y=121705624911623914640817989875157760938714843763893"
[1] "Pred x+y=121705624911623914640817989875157760938714843763893"
\end{verbatim}
\end{tcolorbox}

The rest of this paper is organized as follows: Section 2, `Representative Training Instances,' introduces the two distinct types of training instances used to train the model and explains how these instances facilitate the generalization of addition for numbers of varying digit lengths. Section 3, `Model Setting,' details the technical specifications of the Transformer model used, including its architecture and parameter settings. Section 4, `Generating Tokens from the Trained Model,' describes the autoregressive procedure employed for token generation and demonstrates the model’s application through specific examples. Finally, Section 5, `Final Remarks,' discusses the implications of the findings, reflects on the efficacy of the model, and suggests directions for future research. 

\section{Representative Training Instances}

The vocabulary consists of fourteen tokens: \verb|P S 0 1 2 3 4 5 6 7 8 9 \n C|. Here, \verb|P| represents the padding token, \verb|S| denotes the end of the output, \verb|\n| is used as the initial token for the output, and \verb|C| represents the carry-over token. Although may not be necessary, the carry-over token aids in visualizing the proposed methodology.

To train a Transformer for the addition task with the goal of achieving generalization across arbitrary digit lengths, I propose the construction of two distinct types of training instances: the ones of the first type and the ones of the second type, which are detailed in the following.

\subsection{Instances of the First Type}
For the first type, the input consists of two single-digit numbers and the target is the final solution without using scratchpads. See below some examples.

\begin{tcolorbox}[colback=lightgray, sharp corners, boxrule=0.5mm]
\begin{verbatim}
                 First type of instances: 
             a)  1+2=3    INPUT: 12  TARGET: 3S 
             b)  8+3=11   INPUT: 83  TARGET: 11S 
             c)  6+9=15   INPUT: 69  TARGET: 15S 
             d)  9+9=18   INPUT: 99  TARGET: 18S
\end{verbatim}
\end{tcolorbox}
It is well-known that training a Transformer solely with instances of the first type does not enable generalization to arbitrary digit lengths. This limitation arises because such training does not demonstrate how to apply the carry-over strategy for higher place values. Instances of the first type essentially illustrate the initial step in an addition task. To facilitate the model's learning of the generalization process, additional instances are required. Based on the common experience with manual addition,  numbers comprising two digits can provide such instances, which I refer to as `instances of the second type.'

\subsection{Instances of the Second Type}
In instances of the second type, the input combines the previous outputs, which includes the carry-over information, with the next two single-digit numbers needed to be summed, and the target is the resulting addition with the carry-over. 

For example, \verb|79+14| results in \verb|93|, but we do not usually compute it from left to right. To mimic a common manual calculation, from instances of the first type, the model should learn that \verb|9+4| results in \verb|13|. The model then needs to learn to perform the second stage of addition \verb|7+1|, conditionally based on the result of the first stage. We represent this by the instance of the second type whose input \verb|13C71| indicates that the next addition \verb|7+1| should be performed conditionally on the first stage result \verb|13|. Here, \verb|13| indicates the previous outputs and \verb|71| represents the next two single-digit numbers to be summed. For this case, the target is \verb|9S|, which represents \verb|7+1=8| plus the carry-over from the first operation. Examples of training pairs inputs/targets are provided below.
\begin{tcolorbox}[colback=lightgray, sharp corners, boxrule=0.5mm]
\begin{verbatim}
                   Second type of instances: 
           a)  11+42=53    INPUT: 3C14   TARGET: 5S 
           b)  78+13=91    INPUT: 11C71  TARGET: 9S 
           c)  26+99=125   INPUT: 15C29  TARGET: 12S 
           d)  99+89=188   INPUT: 18C98  TARGET: 18S
\end{verbatim}
\end{tcolorbox}
The model will be able to recognize that any input with only one digit before \verb|C|, such as 3C14, does not have a carry-over, while any input with two digits before \verb|C|, such as 18C98, includes a carry-over to be considered in the operation. This distinction is a straightforward task for the model to handle.

In the training set, instances are randomly generated, with half from each type. However, it may be beneficial to consider increasing the proportion of second-type instances. Inputs and targets are padded to lengths of 5 and 3, respectively, if they fall short of these dimensions. See the following examples to illustrate this approach.
\begin{tcolorbox}[colback=lightgray, sharp corners, boxrule=0.5mm]
\begin{verbatim}
                     Training examples:             
          6+9=15      INPUT: 69PPP  TARGET: 15S 
          78+13=91    INPUT: 11C71  TARGET: 9SP 
          8+3=11      INPUT: 83PPP  TARGET: 11S 
          99+89=188   INPUT: 18C98  TARGET: 18S
          11+42=53    INPUT: 3C14P  TARGET: 5SP 
          9+9=18      INPUT: 99PPP  TARGET: 18S
          26+99=125   INPUT: 15C29  TARGET: 12S
          1+2=3       INPUT: 12PPP  TARGET: 3SP 
\end{verbatim}
\end{tcolorbox}
While it is possible to employ a GPT-like model by concatenating inputs and targets into a single long string and training the model to predict the next token in a typical fashion, this approach is not guaranteed to work effectively and is not adopted in this work. Instead, instances are processed individually without concatenating the inputs and targets. Further details of the model used are described in the following section; see also the R codes in \url{https://github.com/AGPatriota/ALGA-R/}.

\section{Model Setting}

The study employed a decoder-only configuration. All computations were conducted using the statistical software  \cite{R2024}, utilizing the \emph{torch} package.
 The model processes an input $x$ and an initial output $y$ (with the initial token \verb|\n|). These matrices are concatenated vertically to form $z = (x,y)$, a tensor of integers reporting the observed positions in the vocabulary. Subsequently, the tensor undergoes standard embedding procedures to a dimension of 64, complemented by positional encoding. The architecture includes a multi-head attention mechanism with two heads, a feed-forward network (FFN) with a hidden dimension of 256 and two layers. A final linear weight is applied to transform the embedding dimension into the vocabulary dimension. Only the submatrix referring to $y$ is presenting as the output. As the input does not require masking, the attention mask is modified to accomodated this feature; see the Figure \ref{fig1} below.

\begin{figure}[!htp]
\begin{center}
\includegraphics[scale=0.5]{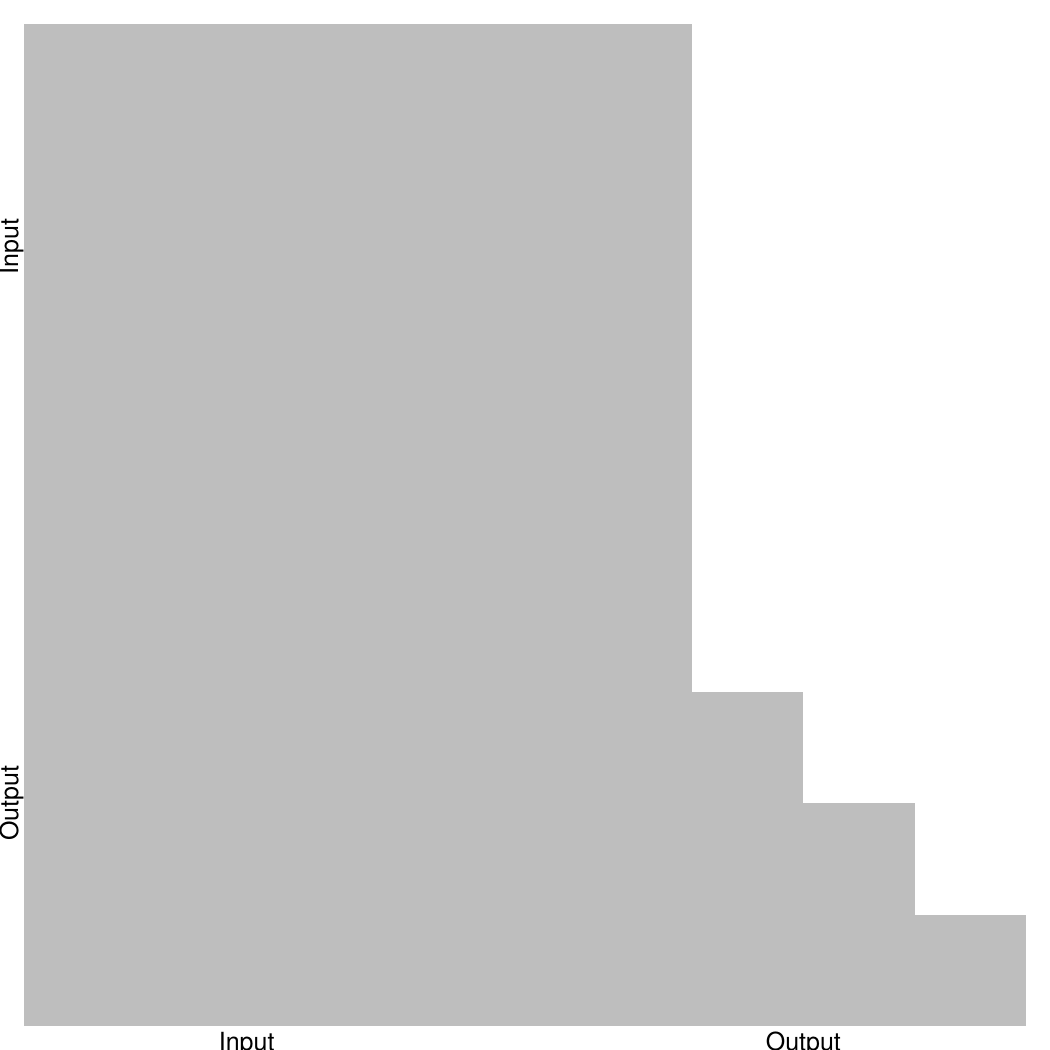}
\end{center}\caption{Masking attention for the concatenated input/output. The gray area represents the unmask region.}\label{fig1}
\end{figure}
The standard Adam optimizer with a weight decay of 0.01 and a learning rate of $5\times10^{-4}$ were adopted. Also, dropout rates of 20\% were applied to both the attention mechanism and the FFN. A batch size of 512 was considered.

\section{Generating Tokens from the Trained Model}

As the training targets contain the intermediate sums (with carry-overs) rather than the final addition, the model should be extensively trained on instances of the first and second type to minimize the risk of missing intermediate predictions during token generation.  The trained model can handle at most two single-digit numbers at a time and the tokens are generated using an autoregressive procedure that starts from the initial output token \verb|\n| and stops the generation when \verb|S| is generated. The outputs attained from a previous input are integrated with the subsequent two single-digit numbers to be summed. As previously discussed, this forms an input of the second type, which contains sufficient information to generate the next token.

To illustrate this process, consider the operation \verb|65785+8765|. Initially, the model processes the input \verb|55|, producing the outputs \verb|10S|. The following input, \verb|10C86|, results in the outputs \verb|15S|, which represents \verb|8+6=14| plus the carry-over from the previous addition. This pattern continues with the input \verb|15C77| leading to the outputs \verb|15S| (also \verb|7+7=14| plus the carry-over), and the input \verb|15C58| yielding \verb|14S| (which is \verb|5+8=13| plus the carry-over). The final input, \verb|14C6|, produces \verb|7S| (which is \verb|6+0=6| plus the carry-over). The overall result is derived by collating the last digits of each block of outputs from last to first, except for the final generation where all digits are considered. In this scenario, \verb|74550| is the resultant sum. Notice that if the last outputs were \verb|17S| rather than \verb|7S|, the final answer would be \verb|174550|. The discussed process is repeated in the following.

\begin{tcolorbox}[colback=lightgray, sharp corners, boxrule=0.5mm]
\begin{verbatim}
      Autoregressive generation to compute: 65785+8765
       1ª) INPUT: 55    generates three OUTPUTS: 10S 
       2ª) INPUT: 10C86 generates three OUTPUTS: 15S 
       3ª) INPUT: 15C77 generates three OUTPUTS: 15S 
       4ª) INPUT: 15C58 generates three OUTPUTS: 14S
       5ª) INPUT: 14C6  generates two   OUTPUTS: 7S         
Read the last digits of each block of outputs from bottom to top
   including all digits of the 5ª line (the last generation)
                      Final answer: 74550
\end{verbatim}
\end{tcolorbox}

\begin{tcolorbox}[colback=lightgray, sharp corners, boxrule=0.5mm]
\begin{verbatim}
      Autoregressive generation to compute: 9582+9261
       1ª) INPUT: 21    generates two   OUTPUTS: 3S 
       2ª) INPUT: 3C86  generates three OUTPUTS: 14S 
       3ª) INPUT: 14C52 generates two   OUTPUTS: 8S 
       4ª) INPUT: 8C99  generates three OUTPUTS: 18S
Read the last digits of each block of outputs from bottom to top
   including all digits of the 4º line (the last generation)
                      Final answer: 18843
\end{verbatim}
\end{tcolorbox}

\subsection{Small experiment}

More than one thousand additions involving numbers with up to one thousand digits each were tested using the trained model, which accurately generated the correct output for {\bf all} tested instances. Additionally, the model was evaluated on some instances with numbers extending to 300 thousand digits, yielding positive results across all tests. However, due to limitations in computational resources, I was unable to conduct these experiments on a larger scale. While it is not feasible to provide an estimate for the accuracy of the procedure for additions of numbers with more than one thousand digits, the results suggest near 100\% accuracy for extremely large numbers.  Anyone interested can reproduce these results by running (or adapting) the R codes in \url{https://github.com/AGPatriota/ALGA-R/}. For specific tests with high-digit numbers, see the function \verb|Testing-Digits.R|.

Despite concerns that the procedure may not generalize as expected, since it does not handle the entire sequence to perform addition, it is important to recognize that humans also do not compute additions of large numbers in a single step or by retaining all digits in  memory. Typically, we add numbers in segments, focusing on two digits at a time, similar to the presented proposal.

\section{Final Remarks}

This paper has explored a novel methodology for training Transformer models to perform addition, emphasizing the need for training instances that are representative of the arithmetic process. I introduced two types of training instances, namely, the ones of the first type, which solely focus on the result of adding two single-digit numbers, and the ones of the second type, which incorporate previous outputs with the carry-over information. This strategy has proven exceptionally effective, enabling the generalization of addition from two-digit numbers to sequences of arbitrary length. The approach demonstrated here, where training data reflect the incremental complexities of the task, could potentially be adapted to train models for more complex arithmetic operations or in tasks where repetitive procedures are applied in long input/output sequences. The tested model flawlessly processed more than one thousand additions involving numbers with up to one thousand digits each, underscoring the efficacy of the method. A notable limitation of this study is that the speed of the learning process was not evaluated, due to a lack of necessary computational resources for conducting such extensive experiments. Future research could focus on exploring the optimal number of instances of the second type required to achieve perfect accuracy across various digit lengths. Also, one could verify what is the minimal information an input of the second type must have to reach perfect accuracy. Furthermore, it might be investigated if it is possible to reduce the autoregressive generations to attain the final answer.

\end{document}